\documentclass{article}

\usepackage{arxiv}

\usepackage[utf8]{inputenc} 
\usepackage[T1]{fontenc}    
\usepackage{hyperref}       
\usepackage{url}            
\usepackage{booktabs}       
\usepackage{amsfonts}       
\usepackage{nicefrac}       
\usepackage{microtype}      
\usepackage{lipsum}
\usepackage{graphicx}
\graphicspath{ {./images/} }

\title{AEGIS: A Semantic GAN and Evidential Learning Framework
for Robust Adversarial Detection in Vision Sensors}

\author{
 Maher Boughdiri \\
  LTCI, Département Informatique et Réseaux, Télécom Paris, Institut Polytechnique de Paris, France
 \\ \texttt{maher.boughdiri@teleecom-paris.fr} \\
   \And
 Mounira Msahli \\
 LTCI, Département Informatique et Réseaux, Télécom Paris, Institut Polytechnique de Paris, France\\
\texttt{mounira.msahli@telecom-paris.fr} \\
  \And
 Albert Bifet \\
 LTCI, Département Informatique et Réseaux, Télécom Paris, Institut Polytechnique de Paris, France\\
  \texttt{albert.bifet@telecom-paris.fr} \\
}
\usepackage{algorithm}
\usepackage{comment}
\usepackage{pifont}
\usepackage{amsmath}
\usepackage{algpseudocode}
\begin{document}

\maketitle
\begin{abstract}
Deep neural networks (DNNs) have shown outstanding performance in visual recognition tasks within vision sensor networks; however, they are still vulnerable to adversarial manipulations and imperceptible perturbations that can lead to erroneous predictions. To address that,  this paper presents \textbf{AEGIS}, a semantic aware and uncertainty guided adversarial detection framework designed for robust image classification in vision sensors pipelines.  At its core, a \textit{SemantiGAN} module functions as a multi class semantic discriminator, identifying and filtering visually inconsistent adversarial inputs before they propagate further in the pipeline. For inputs that pass this stage, a stochastic augmentation process generates test time variations, from which handcrafted instability metrics FlipScore, Prediction Inconsistency, Layerwise Cosine Similarity (early and mid layers), and Entropy are computed. These features are aggregated into a compact five dimensional vector and processed by an Evidential Deep Learning (EDL) classifier, which models output evidence using a Dirichlet distribution to yield both class predictions and calibrated uncertainty estimates. Evaluations on the Tiny ImageNet dataset across six categories clean, FGSM, PGD, patch based, functional, and geometric attacks  demonstrate the effectiveness of AEGIS. 
The proposed framework achieves an AUROC of 92.1\%, an AUPRC of 90.2\%, and an accuracy of 90.7\%, outperforming conventional softmax-based detectors in terms of detection performance, robustness, interpretability, and uncertainty calibration.

\end{abstract}


\section{Introduction}

Deep Neural Networks (DNNs) have become the cornerstone of modern computer vision systems and are increasingly deployed in safety-critical applications such as autonomous driving, facial recognition, medical imaging, and intelligent surveillance \cite{foukalas2025survey}. Despite their remarkable success, DNNs remain highly vulnerable to adversarial attacks, where carefully crafted perturbations can induce incorrect predictions while remaining nearly imperceptible to human observers \cite{vadillo2025adversarial, yerlikaya2026survey}. Such vulnerabilities pose significant safety and security risks. For instance, a manipulated stop sign may be misclassified as a speed limit sign, potentially leading to hazardous decisions in autonomous navigation systems \cite{kurakin2017adversarial}. Furthermore, the transferability of adversarial examples enables successful black-box attacks without requiring direct access to the target model, thereby increasing the practical threat posed by adversarial manipulation \cite{papernot2016limitations}.

Early adversarial attacks primarily focused on pixel-level perturbations constrained by $\ell_p$ norms \cite{tran2022multiple}. Representative methods such as the Fast Gradient Sign Method (FGSM) and Projected Gradient Descent (PGD) generate subtle modifications that preserve the visual appearance of the original image while misleading the classifier \cite{villegas2024evaluating}. More recently, adversarial research has evolved beyond low-level perturbations toward higher-level semantic manipulations \cite{chen2022semantically, zhang2025attacktracer}. These include adversarial patches, geometric transformations, spatial distortions, and functional attacks that alter contextual relationships while preserving human-recognizable content \cite{brown2017adversarial, xiao2018spatially, swain2023spat, yin2021exploiting}. Unlike traditional perturbation-based attacks, semantic adversarial examples exploit weaknesses in contextual reasoning, shape understanding, and texture biases, making them considerably more challenging to detect and defend against \cite{hosseini2018semantic, song2026segtrans}.

To counter these threats, numerous defense mechanisms have been proposed, including input preprocessing, adversarial training, gradient masking, anomaly detection, and uncertainty estimation \cite{madry2018towards, feinman2017detecting, liang2018enhancing}. However, most existing approaches focus on a single aspect of adversarial behavior and often fail to generalize across heterogeneous attack categories. Pixel-level defenses are generally ineffective against patch-based or semantic attacks, while methods designed for $\ell_p$-bounded perturbations remain vulnerable to functional and geometric manipulations \cite{brown2017adversarial, hosseini2018semantic, zhou2018classification}. Moreover, many current defenses provide limited interpretability and uncertainty awareness, reducing their reliability in safety-critical environments where trustworthy decision-making is essential.

The limitations of existing defenses highlight three key requirements for robust adversarial detection. First, semantic consistency analysis is needed to identify attacks that manipulate contextual or structural information while preserving visual realism \cite{liu2025adagat, ye2024mutual}. Second, instability profiling can reveal adversarial behavior by measuring how predictions and internal representations change under benign transformations \cite{swami2025investigating}. Third, uncertainty-aware reasoning is necessary to prevent overconfident predictions and enable the safe rejection of suspicious inputs \cite{li2026two}.

To address these gaps, the main contribution of this work is the proposal of AEGIS, a unified adversarial detection framework that combines semantic reasoning, instability analysis, and evidential uncertainty modeling. It consists of three complementary modules: (1) \textit{SemantiGAN}, a GAN-based semantic discriminator that performs multi-class adversarial filtering; (2) \textit{LAFANet}, an instability profiling module that extracts handcrafted features from test-time augmentations; and (3) \textit{Evidential Deep Learning (EDL)}, which provides calibrated classification and epistemic uncertainty estimation through a Dirichlet-based formulation. By jointly leveraging semantic consistency, prediction instability, and uncertainty awareness, AEGIS offers a comprehensive defense against both low-level perturbation attacks and high-level semantic manipulations.

The remainder of this paper is structured as follows. Section \ref{sec:related} surveys related work in adversarial detection and  defenses approaches. Section \ref{sec:experiments} details the experimental setup and evaluation metrics. Section \ref{sec:threat} defines the adversarial threat model and security objectives. Section \ref{sec:method} presents our full methodology, including SemantiGAN, LAFANet features, and the EDL classifier. Section \ref{sec:results} discusses results, security implications, and ablation studies. Then, we conclude with future directions in Section \ref{sec:conclusion}.
\section{Related Work}
\label{sec:related}
In this section, recent advances in adversarial attack detection and defense strategies in sensors networks are reviewed.

\subsection{ Adversarial Detection Methods}

\subsubsection{Confidence and Statistical-Based Detection}

Early adversarial detection methods relied on confidence-based indicators such as softmax probabilities and predictive entropy, assuming that adversarial perturbations reduce model confidence \cite{liang2018enhancing, feinman2017detecting}. However, adaptive attacks can maintain high confidence while still causing misclassification, significantly limiting the effectiveness of these approaches \cite{carlini2017towards}. To overcome this limitation, statistical methods such as Mahalanobis distance \cite{lee2018simple} and Local Intrinsic Dimensionality (LID) \cite{ma2018characterizing} were introduced to identify deviations in feature-space distributions between clean and adversarial samples. Although these methods achieve promising performance in controlled environments, their effectiveness often deteriorates on deeper architectures and complex high-dimensional datasets.
 \subsubsection{Transformation and Instability-Based Detection}

A growing body of work exploits the instability of adversarial examples under benign input transformations. These methods leverage Test-Time Augmentation (TTA) and stochastic perturbations to assess prediction consistency across multiple transformed views \cite{goyal2020detection, uvdal2025test}. Approaches such as feature squeezing \cite{li2017adversarial}, prediction flipping analysis \cite{zhou2025improving}, and density-aware normalization \cite{xiao2025towards, hasanebrahimi2023density} quantify instability through metrics including entropy, prediction variance, and label inconsistency. More recent studies investigate feature-space divergence, frequency-domain transformations, and stochastic ensemble strategies to improve robustness against adaptive attacks \cite{guesmi2025drift, jia2025adversarial, lao2025test}. Despite their effectiveness, most transformation-based detectors rely on a limited set of instability measures and lack explicit semantic reasoning, reducing their ability to identify semantically aligned adversarial manipulations.

To address the limitations of purely statistical approaches, recent research has incorporated semantic information into adversarial detection. Methods based on attention alignment, scene coherence modeling, and semantic graph reasoning aim to identify violations of semantic consistency rather than low-level perturbation artifacts \cite{sai2024interpretation, cao2025scenetap, gong2023adversarial}. These approaches demonstrate improved robustness against structurally consistent attacks; however, they are often computationally expensive, highly dependent on dataset-specific semantics, and difficult to generalize across diverse attack families.

\subsection{Adversarial Defense Mechanisms}

\subsubsection{GAN-Based Defenses}

Generative Adversarial Networks (GANs) have been widely adopted as defense mechanisms against adversarial attacks. A common strategy consists of projecting adversarial samples back onto the learned data manifold before classification. Defense-GAN \cite{samangouei2018defensegan} exemplifies this approach through iterative latent-space optimization, while anomaly-detection frameworks such as GANomaly \cite{akcay2018ganomaly}, G-VAE \cite{xu2025g}, and RGAnomaly \cite{qian2025rganomaly} leverage reconstruction errors and latent-space deviations to identify abnormal inputs. Although effective in capturing low-level perturbations, reconstruction-based approaches often suffer from high computational overhead, reconstruction artifacts, and reduced effectiveness against semantically coherent attacks such as adversarial patches.

More recently, researchers have explored GAN discriminators as direct adversarial detectors. AdvGAN \cite{xiao2018generating} employs a discriminator to enforce perceptual realism during adversarial generation, while class-conditional GANs improve class-aware discrimination \cite{jiang2020robust}. Advanced architectures such as GEUN \cite{he2025gan} and FSD-GAN \cite{ge2025fsd} integrate multiple discriminative modules to improve robustness against manipulated content. Nevertheless, existing discriminator-based approaches are generally designed for specific attack scenarios and rarely exploit semantic adversarial categories as an explicit detection objective.

\subsubsection{Uncertainty-Aware Learning for Robust Detection}

Conventional neural networks rely on softmax outputs that are often poorly calibrated and prone to overconfidence, particularly when processing adversarial or out-of-distribution samples \cite{fakour2024structured, tuna2022exploiting}. Bayesian approaches such as Monte Carlo Dropout \cite{gal2016dropout} and Deep Ensembles \cite{lakshminarayanan2017simple} estimate predictive uncertainty by approximating posterior distributions over model parameters. Although effective, these methods typically require multiple forward passes and significant computational resources.

Evidential Deep Learning (EDL) \cite{sensoy2018evidential} provides a computationally efficient alternative by modeling classifier outputs as evidence parameters of a Dirichlet distribution. This framework jointly estimates class probabilities and epistemic uncertainty within a single forward pass. Recent studies have demonstrated that EDL improves robustness in adversarial and out-of-distribution detection tasks while providing more reliable uncertainty estimates than conventional confidence-based measures \cite{wang2023adversarial, he2024mutual, holmquist2023evidential}.

\subsection{Research Gaps}

Despite significant progress, existing approaches typically address only isolated aspects of adversarial robustness. Statistical detectors focus on distributional anomalies, transformation-based methods exploit prediction instability, semantic-aware approaches analyze contextual consistency, GAN-based defenses perform purification or specialized discrimination, and uncertainty-aware models provide calibrated confidence estimates. To the best of our knowledge, no prior framework jointly integrates semantic adversarial discrimination, transformation-driven instability profiling, and evidential uncertainty calibration within a unified and computationally efficient architecture. AEGIS addresses this gap by combining SemantiGAN, LAFANet, and Evidential Deep Learning into a single end-to-end framework capable of recognizing six adversarial categories while providing calibrated uncertainty estimates for reliable decision-making.

\section{Threat Model}
\label{sec:threat}
\subsection{Adversary Capabilities and Attack Modalities}
\begin{figure}[h!]
    \centering
    \includegraphics[width=0.8\linewidth]{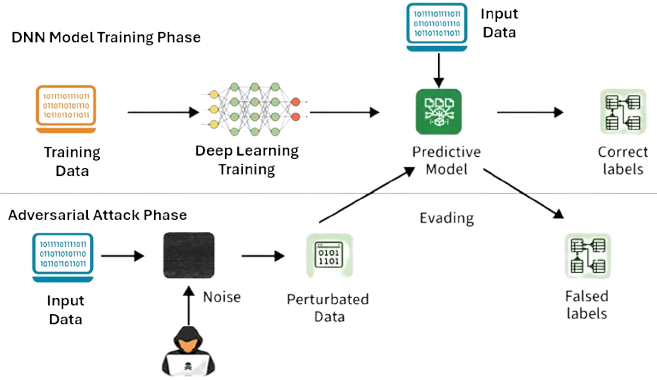}
    \caption{Adversarial Deep Learning Attacks}
    \label{fig:adr}
\end{figure}

Adversarial attacks refer to intentional manipulations of input data designed to mislead a machine learning model into making incorrect predictions while preserving the visual appearance of the input, as illustrated in figure \ref{fig:adr}. These attacks introduce small, often imperceptible perturbations that exploit weaknesses in the model’s decision process. As a result, adversarial perturbations pose a persistent security risk to deep learning-based vision pipelines, particularly in real-time deployments where high-confidence misclassifications can trigger safety-critical failures \cite{szegedy2014intriguing, goodfellow2015explaining}. Such perturbations are crafted to alter the model’s output without changing the human-perceived semantics of the image, thereby undermining the trustworthiness of DNN-enabled decision systems.

Adversarial attacks can be categorized based on the attacker’s level of knowledge into three settings: \textit{white-box} \cite{liu2023gradient}, \textit{black-box} \cite{zhu2024review}, and \textit{gray-box}  \cite{costa2026lisard}. In this work, a \textit{white-box} adversary is considered, assuming complete access to the target model’s architecture, parameters, and gradients. This level of access enables the use of loss-driven backpropagation to construct adversarial perturbations that maximize the probability of misclassifications. Such a setting represents a strong, worst-case scenario and is widely adopted to evaluate the upper bounds of model robustness.

The adversary’s objective is to generate perturbed inputs that induce incorrect or targeted predictions while preserving perceptual similarity, ensuring that the modifications remain imperceptible to human observers. These perturbations are typically constrained under a bounded norm (e.g., $\ell_p$), or designed to maintain semantic plausibility in more advanced attack scenarios.
Under this threat model, the following classes of adversarial attacks are considered:  
\begin{itemize}
    \item $\ell_p$-bounded gradient-based attacks: Methods such as Fast Gradient Sign Method (FGSM)  and Projected Gradient Descent (PGD) \cite{ali2025white, rub2026pgd} exploit gradient information to compute adversarial perturbations that maximize the loss function with respect to the input. These perturbations are constrained within a predefined norm ball (e.g., $\ell_\infty$, $\ell_2$), ensuring imperceptibility while effectively pushing samples across decision boundaries. PGD, as an iterative extension of FGSM, is widely regarded as a first-order worst-case adversary due to its stronger optimization capability.

    \item Patch-based attacks: These attacks introduce localized, high-impact perturbations confined to a small region of the input image \cite{brown2017adversarial, lian2022benchmarking}. Unlike global perturbations, adversarial patches are optimized to dominate salient features and attention mechanisms, often overriding the original content. Their spatial locality and robustness make them particularly effective in both digital and physical settings.

    \item Functional (semantic) attacks: Rather than relying on low-level noise, these attacks manipulate high-level attributes such as object texture, material properties, or contextual relationships \cite{joshi2019semantic}. The resulting samples remain visually plausible to humans but induce semantic inconsistencies that mislead the model, revealing its reliance on non-robust or spurious correlations.

    \item Geometric transformation attacks: These attacks apply spatial transformations, including rotations, translations, scaling, and non-linear warpings \cite{chen2025g}. Such transformations exploit the limited invariance of convolutional neural networks, leading to significant changes in internal feature representations despite minimal perceptual distortion.

    \item Transferable adversarial examples: Although commonly associated with black-box settings, transferability can also be studied under white-box conditions by crafting perturbations that generalize across models. These adversarial examples exploit shared decision boundaries and feature representations, highlighting systemic vulnerabilities in deep learning models \cite{papernot2016transferability}.

    \item Hybrid/zero-day attacks: These attacks combine multiple perturbation strategies (e.g., gradient-based noise with geometric distortion or semantic manipulation) to produce more complex and evasive adversarial examples. Such compositions often fall outside the training distribution and challenge detection mechanisms, representing realistic and previously unseen (zero-day) threat scenarios \cite{carlini2017towards}.
\end{itemize}

\subsection{Security Landscape and Design Constraints}

Beyond conventional adversarial attacks, the deployment of vision-based AI systems in real-world environments introduces additional security and operational challenges that must be considered during the design of adversarial defense mechanisms. A  critical challenge is adversarially induced overconfidence, where adversarial inputs not only trigger incorrect predictions but also produce artificially high confidence scores, concealing the model's uncertainty and potentially misleading downstream decision-making processes \cite{hein2019relu}. Such behavior is especially problematic in safety-critical applications, where confidence estimates are often used to trigger automated actions or risk mitigation procedures.

An effective adversarial defense must simultaneously satisfy three core requirements: robustness against a wide range of adversarial manipulations, compliance with real-time inference constraints, and the ability to provide transparent and interpretable decision-making. AEGIS is designed to address these requirements through a modular architecture comprising three complementary components. First, SemantiGAN enables early rejection of semantically inconsistent or out-of-distribution inputs by modeling class-level semantic coherence. Second, LAFANet leverages instability metrics derived from lightweight test-time augmentations to expose hidden sensitivity to adversarial perturbations, including gradient-based and geometric distortions. Third, Evidential Deep Learning (EDL) provides calibrated predictions through principled uncertainty estimation, enabling reliable rejection of ambiguous or adversarially influenced inputs. Each module in AEGIS is explicitly aligned with a distinct adversarial capability: SemantiGAN targets semantic and patch-based attacks by filtering visually inconsistent inputs, LAFANet captures vulnerability to gradient and geometric perturbations through prediction instability, and EDL addresses adversarially induced overconfidence by quantifying epistemic uncertainty. This unified design mitigates a broad spectrum of modeled threats; however, residual risks remain in more complex settings such as physical-world robustness and adaptive zero-day attacks, which are further discussed in Section~\ref{sec:results}. 

\section{AEGIS  Framework}
\label{sec:method}

\begin{figure}[h!]

    \centering
    \includegraphics[width=1.1\linewidth]{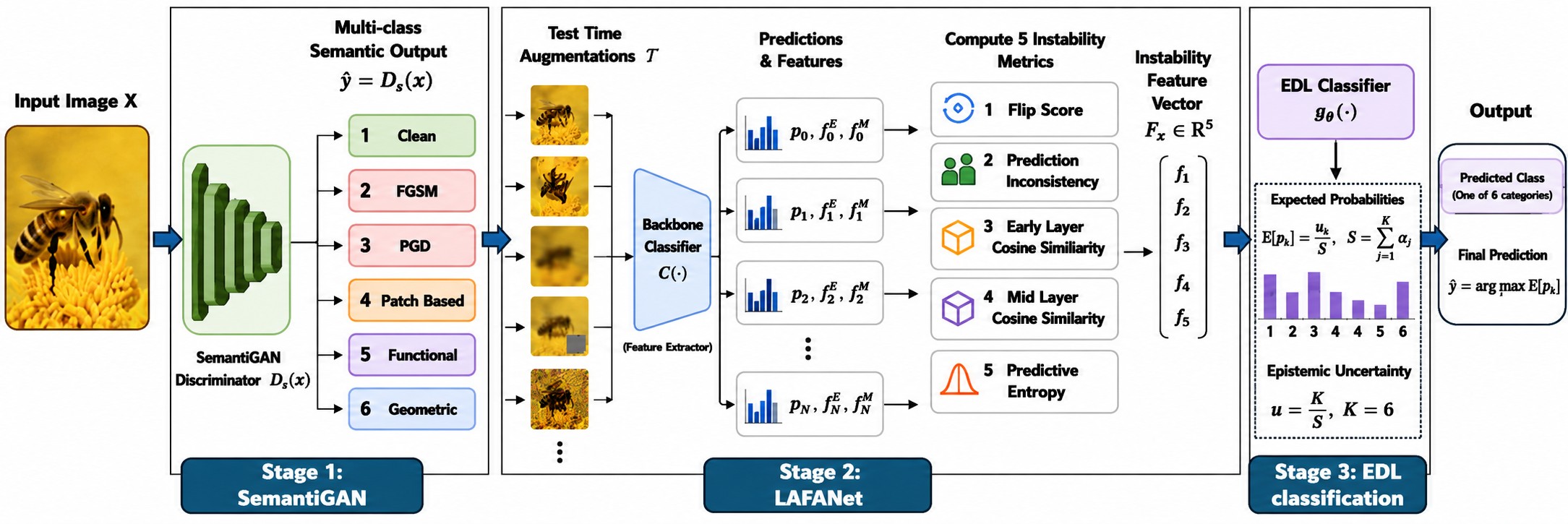}
    \caption{Overview of AEGIS adversarial detection framework}
    \label{fig:safe}

\end{figure}

AEGIS is a modular adversarial detection and classification framework that integrates three complementary components: (i) SemantiGAN   a multi class semantic discriminator based on a generative adversarial network (GAN), (ii) LAFANet   an instability profiling module extracting handcrafted features from test time augmentations (TTA), and (iii) Evidential Deep Learning (EDL)   a calibrated classifier that models epistemic uncertainty. The overall pipeline, illustrated in Figure \ref{fig:safe}, processes every input image through all three stages, ensuring that both clean and adversarial inputs are consistently analyzed for final classification across six categories. In the first stage, SemantiGAN performs multi-class semantic discrimination by recognizing different categories of adversarial perturbations and providing a coarse semantic characterization of the input. However, semantic discrimination alone is insufficient to reliably determine whether an image is truly adversarial, as different attack types may exhibit similar semantic characteristics. To address this limitation, the second stage employs LAFANet, which analyzes prediction instability under stochastic, label-preserving augmentations and extracts a set of handcrafted instability features. Finally, these features are provided to the EDL classifier, which produces the final adversarial classification while quantifying epistemic uncertainty. This uncertainty-aware decision process improves the reliability, robustness, and interpretability of the framework, particularly when confronted with ambiguous, adversarial, or out-of-distribution inputs.

\subsection{Framework stages} 
The AEGIS framework is composed of three stages: (1) SemantiGAN, a multi-class semantic discriminator; (2) Augmentation-Driven Instability Profiling (LAFANet); and (3) Evidential Deep Learning (EDL) Classification.

\subsection*{Stage 1: SemantiGAN - Multi Class Semantic Discriminator}
The first stage employs a semantic GAN discriminator
$D_s(\cdot)$ trained to classify an input image $x$ into one of $K=6$ semantic categories:
\begin{equation}
\hat{\mathbf{y}} = D_s(x) = \mathrm{softmax}(f_{\theta}(x)) \in \mathbb{R}^6,
\label{eq:semantigan_output}
\end{equation}
where $f_{\theta}$ denotes the discriminator’s feature extractor. The six categories include: \textit{clean}, FGSM, PGD, patch based, functional, and geometric adversarial examples.  

Unlike traditional GAN discriminators optimized for binary real/fake detection, SemantiGAN is directly trained for multi class adversarial recognition, enabling it to learn high level semantic differences between diverse attack types and clean inputs.  

The output from SemantiGAN serves as a coarse semantic cue that complements the downstream instability analysis.

\subsection*{Stage 2: Augmentation Driven Instability Profiling (LAFANet)}

Following semantic classification, the image is passed to LAFANet, which evaluates prediction stability under stochastic, label-preserving transformations. These transformations are defined by a set of lightweight operators $\mathcal{T}$ such as horizontal flips, affine warps, brightness adjustments, motion blur, cutout, and JPEG compression.

Given an input image $x$, $N$ stochastic augmentations are generated as:
\begin{equation}
\{x_i\}_{i=1}^{N} = \mathcal{T}(x)
\end{equation}

Each image (original and augmented) is processed by a backbone classifier $C(\cdot)$ producing class probability distributions:
\begin{equation}
p_0 = C(x), \quad p_i = C(x_i), \quad p_i \in \mathbb{R}^K
\end{equation}

The corresponding predicted labels are:
\begin{equation}
\hat{y}_0 = \arg\max p_0, \quad \hat{y}_i = \arg\max p_i
\end{equation}

From these outputs, five instability metrics are computed. These metrics are defined as follows.

\textbf{Flip Score:} Measures how often the predicted class changes under a stochastic augmentation $\mathcal{T}_i$ . Mainly, its a binary indicator of label change under horizontal flip \cite{girardi2024post}.
\begin{equation}. 
FlipScore(x) = \frac{1}{N} \sum_{i=1}^{N} \mathbb{1}\left[\arg\max p_i \neq \arg\max p_0\right]
\end{equation}

\textbf{Prediction Inconsistency:} Measures the dispersion of predicted labels across augmentations.  It is the fraction of augmented views disagreeing with the majority prediction \cite{garg2023rlsbench}.  
\begin{equation}
Inconsistency(x) = 1 - \frac{\max_c \sum_{i=1}^{N} \mathbb{1}[\hat{y}_i = c]}{N}
\end{equation}

\textbf{Feature Stability:} Measures consistency of learned representations under augmentations: Cosine similarity between early-layer features of each view and the mean early-layer feature \cite{xu2025good};  and  the Cosine similarity between mid layer features of each view and the mean mid layer feature \cite{esposito2024bridging}.
\begin{equation}
EarlyCosSim(x) = \cos\left(f^E(x), \bar{f}^E\right), \quad
MidCosSim(x) = \cos\left(f^M(x), \bar{f}^M\right)
\end{equation}

where:
\[
\bar{f}^E = \frac{1}{N}\sum_{i=1}^{N} f^E(x_i), \quad
\bar{f}^M = \frac{1}{N}\sum_{i=1}^{N} f^M(x_i)
\]
$\bar{f}^E$ and $\bar{f}^M$ denote mean early and mid layer features over augmentations.

\textbf{Predictive Entropy:} Measures uncertainty in the predicted class distribution across augmentations \cite{ye2023improving}.
\begin{equation}
Entropy(x) = \frac{1}{N} \sum_{i=1}^{N} \left(-\sum_{j=1}^{K} p_{ij} \log p_{ij} \right)
\end{equation}

where:$p_{ij}$ denotes the predicted probability of class $j$ for augmentation $x_i$.\\
The resulting 5D instability feature vector is:
\begin{equation}
F_x = [
FlipScore,\ Inconsistency,\ EarlyCosSim,\ MidCosSim,\ Entropy
]
\end{equation}

\subsection*{Stage 3: Evidential Deep Learning (EDL) Classification}
The final stage of AEGIS performs uncertainty-aware adversarial classification using an Evidential Deep Learning (EDL) classifier \cite{sensoy2018evidential}. 
Unlike conventional softmax classifiers that produce deterministic confidence scores, EDL models predictive uncertainty by interpreting network outputs as evidence supporting different hypotheses. This enables the framework to jointly estimate both class probabilities and the confidence associated with those predictions.

In this stage, the 5D feature vector $F_x$ is provided as input to the EDL classifier, which produces non negative evidence $\mathbf{e} \in \mathbb{R}^K_{\ge 0}$. The corresponding Dirichlet parameters are:
\begin{equation}
\boldsymbol{\alpha} = \mathbf{e} + 1,
\label{eq:dirichlet_params}
\end{equation}
and the expected probability for each class is:
\begin{equation}
\mathbb{E}[p_i] = \frac{\alpha_i}{S}, \quad S = \sum_{j=1}^K \alpha_j.
\label{eq:expected_prob}
\end{equation}
Epistemic uncertainty is computed as:
\begin{equation}
u = \frac{K}{S}.
\label{eq:uncertainty}
\end{equation}
Here, epistemic uncertainty refers to uncertainty arising from lack of model knowledge, as opposed to aleatoric uncertainty, which is caused by inherent noise in the data. In adversarial detection, epistemic uncertainty is especially important because it reflects when the model encounters inputs that lie outside the training distribution.

The EDL classifier provides the final six-class prediction along with an uncertainty score. High uncertainty values can be flagged for human review or risk-aware decision-making in safety-critical deployments.

The EDL classifier transforms the raw evidence into a Dirichlet distribution that represents both the most likely class and how confident the model is about that choice. A high evidence value indicates strong confidence, while low evidence reflects greater uncertainty. This provides a principled way to reject inputs that may be adversarial or ambiguous.

In this stage, the 5-dimensional instability feature vector \(F_x\), extracted by LAFANet, is provided as input to the EDL classifier:
\begin{equation}
\mathbf{e}=g_\phi(F_x)\in\mathbb{R}_{\ge0}^{K},
\label{eq:evidence}
\end{equation}
where:
\begin{itemize}
    \item \(g_\phi(\cdot)\) denotes the evidential classifier,
    \item \(K=6\) corresponds to the adversarial categories,
    \item and each component \(e_i\) represents the amount of evidence supporting class \(i\).
\end{itemize}

The evidence vector is then transformed into Dirichlet distribution  parameters:
\begin{equation}
\boldsymbol{\alpha}=\mathbf{e}+1,
\label{eq:dirichlet_params}
\end{equation}
with:
\[
\alpha_i=e_i+1.
\]

The resulting Dirichlet distribution:
\begin{equation}
\mathrm{Dir}(\mathbf{p}\mid\boldsymbol{\alpha}),
\end{equation}
provides a probabilistic representation over class predictions, enabling the model to jointly capture belief strength and predictive uncertainty.

The expected probability for each class is computed as:
\begin{equation}
\mathbb{E}[p_i]
=
\frac{\alpha_i}{S},
\qquad
S=\sum_{j=1}^{K}\alpha_j,
\label{eq:expected_prob}
\end{equation}
where \(S\) denotes the total accumulated evidence.

Epistemic uncertainty is estimated through:
\begin{equation}
u=\frac{K}{S},
\label{eq:uncertainty}
\end{equation}
where higher uncertainty values correspond to lower accumulated evidence.

Here, epistemic uncertainty refers to uncertainty arising from insufficient model knowledge, in contrast to aleatoric uncertainty, which originates from inherent noise in the observations. In adversarial detection, epistemic uncertainty is particularly important because adversarial or out-of-distribution inputs often lie outside the learned training distribution, leading to reduced evidence accumulation and consequently higher uncertainty scores.

The final adversarial prediction is obtained as:
\begin{equation}
\hat{y}=\arg\max_i \mathbb{E}[p_i].
\end{equation}

The EDL classifier therefore transforms the extracted instability features into a Dirichlet-based evidential representation that simultaneously models the most likely adversarial category and the confidence associated with that prediction. High evidence values indicate strong support for a specific class, whereas low evidence reflects ambiguity or insufficient model confidence. This provides a principled mechanism for identifying unreliable predictions and rejecting potentially adversarial or ambiguous inputs. Consequently, samples associated with high uncertainty may be flagged for human inspection or risk-aware decision-making in safety-critical deployments such as autonomous driving, intelligent surveillance, industrial robotics, and medical imaging systems.

\subsection{Design Rationale and Architectural Motivation}
The AEGIS framework is designed according to a progressive adversarial analysis strategy that combines semantic reasoning, instability profiling, and uncertainty-aware decision making. Although SemantiGAN, LAFANet, and Evidential Deep Learning (EDL) can each operate independently as adversarial detection mechanisms, their sequential integration enables a more comprehensive characterization of adversarial behavior across semantic, statistical, and probabilistic dimensions.
\begin{algorithm}[htbp]
\caption{AEGIS: End to End Six Class Adversarial Classification}
\label{alg:AEGIS_class}
\textbf{Input}: Image $x$\\
\textbf{Output}: (Predicted class $c^*$, Uncertainty $u$)\\
\textbf{1:} $\hat{\mathbf{y}}_{\text{sem}} \leftarrow D_s(x)$ \textit{SemantiGAN coarse prediction} \\
\textbf{2:} Apply $N$ stochastic augmentations to $x$;\\ 
\textbf{3: For} each $x_i$, extract $\hat{y}_i$, $f^E(x_i)$, $f^M(x_i)$ \\
\textbf{4:} Compute FlipScore, Inconsistency, EarlyCosSim, MidCosSim, Entropy; form $F_x$. \\
\textbf{5:} $(\mathbf{e}, u) \leftarrow \mathrm{EDL}(F_x)$; compute $\boldsymbol{\alpha}$ and $\mathbb{E}[p_i]$. \\
\textbf{6:} $c^* \leftarrow \arg\max \mathbb{E}[p_i]$; \\\Return $(c^*, u)$.
\end{algorithm}
As presented in algorithm \ref{alg:AEGIS_class}, the first stage employs \textit{SemantiGAN} as a semantic discriminator that analyzes the high-level structure and contextual consistency of the input image. This stage serves as an initial filtering mechanism by identifying semantic discrepancies associated with adversarial manipulations. Such a design is particularly effective against patch-based, functional, and geometric attacks that alter semantic representations while preserving visual plausibility. Performing semantic analysis first is computationally efficient and provides coarse adversarial priors that guide subsequent stages.
The second stage utilizes \textit{LAFANet} to evaluate the stability of model predictions under a set of lightweight stochastic augmentations. While semantically plausible adversarial examples may evade the first stage, they frequently exhibit unstable behavior when subjected to benign transformations. LAFANet captures this phenomenon through handcrafted instability metrics that quantify prediction consistency, feature-space coherence, and uncertainty variations. Compared with computationally intensive approaches such as Mahalanobis distance estimation, Local Intrinsic Dimensionality (LID), or ensemble-based detectors, LAFANet offers a lightweight and architecture-agnostic mechanism for exposing adversarial sensitivity.
The final stage employs an \textit{Evidential Deep Learning (EDL)} classifier to transform the extracted instability representation into class predictions accompanied by calibrated uncertainty estimates. 

Unlike conventional softmax classifiers, which often produce overconfident predictions for adversarial or out-of-distribution samples, EDL explicitly models evidence and epistemic uncertainty through a Dirichlet distribution. Consequently, the framework can distinguish between confident adversarial classifications and ambiguous inputs for which insufficient evidence exists.

\section{Experimental Setup}
\label{sec:experiments}
\subsection{System Configuration}
All experiments were conducted on a high-performance workstation running Windows 11, equipped with an Intel64 processor (Family 6, Model 183) featuring 20 physical cores and 28 logical threads, 32~GB of RAM, and an NVIDIA RTX 2000 Ada Generation Laptop GPU with CUDA~12.8 support. The AEGIS framework was implemented in Python~3.10 using PyTorch~2.7.0+cu128 and TorchVision~0.16.0 for model development and training. Data augmentation and image transformation pipelines were implemented using Albumentations~1.3, while NumPy was employed for instability feature extraction and vector construction. All stages of the framework, including SemantiGAN training, LAFANet instability profiling, and Evidential Deep Learning (EDL) classification, were executed locally using single-GPU acceleration. 
For SemantiGAN, both the generator and discriminator adopted a multi-path convolutional architecture, and the discriminator was trained using categorical cross-entropy loss with the Adam optimizer (learning rate $1 \times 10^{-4}$, $\beta_1=0.5$, $\beta_2=0.999$) for 100 epochs with a batch size of 128. To improve generalization, dropout with a rate of 0.3 and weight decay of $1 \times 10^{-5}$ were applied. The EDL classifier was implemented as a three-layer multilayer perceptron (MLP) with hidden dimensions of 128, 64, and 32 neurons and ReLU activation functions. Training was performed using the Adam optimizer with a learning rate of $5 \times 10^{-4}$, a batch size of 64, and 80 epochs. The evidential loss function combined mean squared error with Kullback--Leibler (KL) divergence regularization to learn calibrated uncertainty estimates. This experimental configuration provided sufficient computational resources for adversarial example generation, large-scale augmentation processing, semantic adversarial classification, instability profiling, and uncertainty-aware evaluation across all six semantic classes.


\subsection{Dataset }
To evaluate the effectiveness of the AEGIS framework, we use the Tiny ImageNet-200 dataset, a widely adopted benchmark for image classification and adversarial learning. Tiny ImageNet consists of 200 object categories selected from the larger ImageNet hierarchy, with each class containing 500 training images, 50 validation images, and 50 test images. All images are resized to a resolution of $64 \times 64$ pixels, preserving the semantic diversity of ImageNet while remaining computationally efficient.

For our experiments, we retain the original class distribution while repurposing the validation and test splits to include adversarial variants. The training split is primarily composed of clean images and is used to train the backbone classifier, LAFANet, and the Evidential Deep Learning (EDL) classifier. However, the SemantiGAN module requires supervision across all six semantic categories (\emph{clean}, FGSM, PGD, patch-based, functional, and geometric attacks). Therefore, adversarial examples were generated from the training split and used exclusively to train SemantiGAN.

In contrast, both LAFANet and the EDL classifier were trained solely on clean images. During training, LAFANet learns the characteristic stability patterns of benign samples under stochastic augmentations, while the EDL module learns to map the resulting instability representations to calibrated evidential outputs. This design ensures a clear separation between semantic adversarial recognition and instability-based uncertainty modeling. Specifically, SemantiGAN is trained to recognize adversarial attack categories, whereas LAFANet and EDL operate under a clean-data training regime and detect adversarial behavior through deviations from normal prediction and feature stability patterns.

\begin{table*}[h]
\centering
\caption{Dataset partitioning for AEGIS training and evaluation.}
\label{tab:data_partition}
\begin{tabular}{lccc}
\hline
\textbf{Split} & \textbf{Clean} & \textbf{Adversarial (5 types)} & \textbf{Usage} \\ \hline
Train &100k & 100k (generated) & SemantiGAN \\
Validation & 10k & 10k & Hyperparameter tuning \\
Test & 10k & 50k (balanced across 5 attacks) & Final evaluation \\
\hline
\end{tabular}
\end{table*}

The validation set is used for hyperparameter tuning and early stopping. The test set is augmented with multiple adversarial versions of the original images, enabling robust evaluation of multi-class detection performance across clean and perturbed data. To ensure a balanced evaluation, we construct a six-class test set where each original image is associated with five corresponding adversarial variants generated using different attack methods. The resulting test set includes clean samples and adversarial examples from FGSM, PGD, patch-based, functional, and geometric attacks. The adversarial portion of the test set contains 50,000 samples, evenly partitioned with 10,000 samples per attack type. Adversarial examples for \emph{evaluation} were generated exclusively from the test set to avoid train–test contamination, whereas adversarial examples for \emph{training} SemantiGAN were synthesized from the training split.

\subsection{Augmentation Protocol and Instability Metrics}
\label{sec:exp_augmentation}

Stage~2 of AEGIS employs a lightweight, label-preserving augmentation pipeline to induce controlled variability in test-time predictions. The augmentation set, implemented using the Albumentations library, includes spatial transformations such as horizontal flips and random affine transformations, photometric perturbations including brightness jitter and motion blur, content masking via cutout with random mask size and position, and compression noise through JPEG artifact injection.

Each input image generates $N=10$ augmented views $\{x_i\}_{i=1}^{N}$, which are processed through a fixed backbone network to obtain predicted labels $\{\hat{y}_i\}$ along with early- and mid-level representations $\{f^E(x_i), f^M(x_i)\}$.

From these outputs, the five instability metrics are computed to form the LAFANet feature vector $F_x$, capturing complementary aspects of prediction variability, feature-space consistency, and model uncertainty. These handcrafted features quantify semantic and representational instability induced by adversarial perturbations while maintaining low computational overhead, making the approach both efficient and effective for robustness analysis.


\subsection{Evaluation Metrics and Baselines}

We employ a diverse set of quantitative metrics to comprehensively evaluate detection accuracy, robustness, and trust calibration. The primary classification metric is overall accuracy across the six classes, which includes clean and all five adversarial types. In addition, we report macro averaged precision, recall, and F1 score to account for potential class imbalance and to capture fine grained detection fidelity.

To assess binary detection capability distinguishing clean from adversarial we compute the Area Under the Receiver Operating Characteristic curve (AUROC), which provides a threshold independent measure of separability. We also analyze per class AUROC to determine the model’s sensitivity to specific attack types.

For uncertainty evaluation, we report the average epistemic uncertainty score produced by the evidential deep learning model for clean and adversarial samples. We further compute the Expected Calibration Error (ECE) to quantify the alignment between predicted confidence and true likelihood. These metrics provide insight into the trustworthiness of AEGIS’s decisions, particularly under adversarial threat.


\section{Results Analysis and Discussions}
\label{sec:results}

This section presents a comprehensive evaluation of the AEGIS framework under diverse adversarial conditions. Performance is assessed using standard classification and detection metrics, including AUROC, AUPRC, F1 score, and accuracy, together with analyses of robustness, uncertainty calibration, and generalization. The objective is to evaluate not only the detection capability of AEGIS but also its ability to maintain reliable performance across heterogeneous adversarial threat models.

\subsection{Quantitative Detection Performance}

Table~\ref{tab:quantitative_metrics} summarizes the detection performance of AEGIS across five representative adversarial attack categories. Overall, the framework achieves an average AUROC of 91.8\%, an average AUPRC of 89.98\%, an F1 score of 89.32\%, and an overall classification accuracy of 90.12\%, demonstrating strong discrimination between clean and adversarial samples across multiple attack families.
\begin{table*}[ht]
\centering
\caption{Detection performance of AEGIS across adversarial attack types using AUROC, AUPRC, F1 score, and Accuracy metrics.}
\label{tab:quantitative_metrics}
\begin{tabular}{lcccc}
\toprule
\textbf{Attack Type} & \textbf{AUROC (\%)} & \textbf{AUPRC (\%)} & \textbf{F1 Score (\%)} & \textbf{Accuracy (\%)} \\
\midrule
FGSM        & 96.1 & 94.8 & 94.7 & 95.3 \\
PGD         & 94.2 & 92.5 & 92.3 & 93.0 \\
Patch       & 91.0 & 88.9 & 88.5 & 89.2 \\
Geometric   & 88.3 & 86.2 & 85.1 & 86.0 \\
Functional  & 89.4 & 87.5 & 86.0 & 87.1 \\
\midrule
\textbf{Average} & \textbf{91.8} & \textbf{89.98} & \textbf{89.32} & \textbf{90.12} \\
\bottomrule
\end{tabular}
\end{table*}

Among the evaluated attacks, FGSM exhibits the highest detection performance, achieving an AUROC of 96.1\% and an accuracy of 95.3\%. This result is expected, as single-step gradient perturbations often introduce characteristic instability patterns that are effectively captured by the LAFANet features. PGD attacks remain highly detectable, achieving an AUROC of 94.2\%, confirming that the proposed instability profiling mechanism remains effective against stronger iterative gradient-based attacks.

More challenging attack classes include patch-based, functional, and geometric perturbations. These attacks preserve much of the original image structure while manipulating semantic content or spatial relationships, making them inherently more difficult to detect. Nevertheless, AEGIS maintains AUROC values above 88\% across all three categories, demonstrating the effectiveness of the combined semantic-discrimination and instability-analysis strategy. In particular, the performance on geometric attacks (88.3\% AUROC) highlights the contribution of SemantiGAN in capturing semantic inconsistencies that may not be evident through prediction confidence alone.

\subsection{Cross-Dataset Generalization}

To evaluate the scalability and transferability of AEGIS beyond Tiny ImageNet, we conducted preliminary experiments on ImageNet-128. As shown in Table~\ref{tab:imagenet128}, the framework maintains strong performance despite the increased image resolution and dataset complexity. Detection accuracy decreases only marginally from 90.1\% to 88.3\%, while AUROC remains high at 89.7\%. Furthermore, the Expected Calibration Error (ECE) increases only slightly from 1.6\% to 2.2\%, indicating that the EDL module preserves well-calibrated uncertainty estimates under more challenging conditions.

These results suggest that the proposed combination of semantic discrimination, instability profiling, and evidential uncertainty modeling generalizes effectively across datasets and resolutions, supporting its applicability to real-world computer vision systems operating in dynamic threat environments.

\begin{table*}[h]
\centering
\caption{Preliminary AEGIS results on ImageNet-128.}
\label{tab:imagenet128}
\begin{tabular}{lcc}
\hline
\textbf{Metric} & \textbf{Tiny ImageNet (64$\times$64)} & \textbf{ImageNet-128 (128$\times$128)} \\
\hline
Accuracy (\%) & 90.1 & 88.3 \\
AUROC (\%)    & 91.8 & 89.7 \\
ECE (\%)      & 1.6  & 2.2 \\
\hline
\end{tabular}
\end{table*}
\subsection{Discriminator Output and Semantic Filtering}

As shown in Table~\ref{tab:semantic_filtering}, the SemantiGAN discriminator achieves high precision and recall in identifying adversarial samples while maintaining a low false rejection rate on clean inputs. By acting as an initial semantic filtering stage, it effectively reduces the burden on downstream modules and improves overall decision reliability. Its integration into the AEGIS pipeline increases downstream classification accuracy by more than 4.5\%, highlighting its effectiveness as a semantic gatekeeper for adversarial robustness.

\begin{table*}[ht]
\centering
\caption{Performance of the SemantiGAN discriminator as a semantic filtering module.}
\label{tab:semantic_filtering}
\begin{tabular}{lcc}
\toprule
\textbf{Metric} & \textbf{Value (\%)} & \textbf{Remarks} \\
\midrule
Precision (Adversarial)          & 93.4 & High confidence in adversarial detection \\
Recall (Adversarial)             & 89.6 & Strong detection coverage across attack types \\
False Rejection Rate (Clean)     & 3.8  & Low misclassification of benign inputs \\
Classifier Accuracy (w/o Filter) & 86.2 & Baseline classifier without semantic filtering \\
Classifier Accuracy (w/ Filter)  & 90.7 & Improved reliability with SemantiGAN \\
Optimal Threshold                & 0.72 & Selected for best precision–recall trade-off \\
\bottomrule
\end{tabular}
\end{table*}

\subsection{Uncertainty and Out-of-Distribution Detection}

The Evidential Deep Learning (EDL) classifier within the AEGIS pipeline is evaluated for its ability to quantify uncertainty and produce calibrated predictions under adversarial perturbations. Unlike conventional softmax-based classifiers that tend to produce overconfident outputs, the proposed model estimates a Dirichlet distribution over class probabilities, enabling the detection of ambiguous or malicious inputs through reduced evidence accumulation and increased predictive entropy.

To evaluate this behavior, we report standard calibration and uncertainty metrics, including the Brier score, which measures the mean squared error between predicted confidence and ground-truth labels, and the Expected Calibration Error (ECE), which quantifies the misalignment between predicted confidence and empirical accuracy. In addition, predictive entropy is used as an indicator of uncertainty, where higher values correspond to less confident predictions. These metrics are computed separately for clean samples and each adversarial category, including FGSM, PGD, patch-based, geometric, and functional attacks.

As shown in Table~\ref{tab:uncertainty_metrics}, the classifier exhibits low uncertainty and strong calibration on clean data, while adversarial inputs consistently induce higher entropy and increased calibration error. These results demonstrate the effectiveness of the evidential formulation in distinguishing between epistemic uncertainty caused by adversarial perturbations and normal predictive variability.

\begin{table*}[ht]
\centering
\caption{Uncertainty-based performance of the Evidential Deep Classifier across clean and adversarial samples.}
\label{tab:uncertainty_metrics}
\begin{tabular}{lccc}
\toprule
\textbf{Input Type} & \textbf{Brier Score} $\downarrow$ & \textbf{ECE (\%)} $\downarrow$ & \textbf{Entropy} $\uparrow$ \\
\midrule
Clean Samples   & 0.087 & 1.8 & 0.23 \\
FGSM            & 0.122 & 5.4 & 1.12 \\
PGD             & 0.134 & 6.1 & 1.45 \\
Patch-based     & 0.118 & 5.9 & 1.33 \\
Geometric       & 0.129 & 6.7 & 1.38 \\
Functional      & 0.131 & 6.5 & 1.41 \\
\bottomrule
\end{tabular}
\end{table*}


\subsection{Visualizations}

\begin{figure}[h!]
    \centering
    \includegraphics[width=1\linewidth]{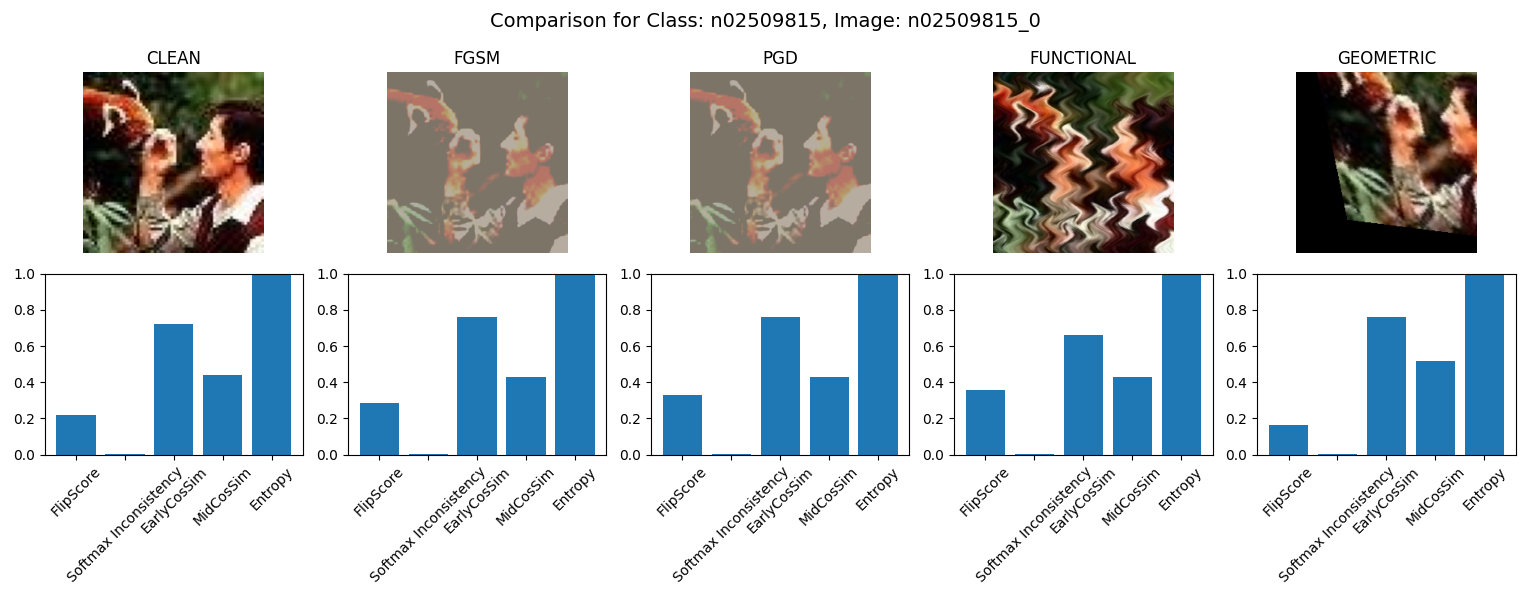}
    \caption{Qualitative comparison of instability metrics for clean and adversarial samples (example 1).}
    \label{fig:fig1_withoutpatch}
\end{figure}

\begin{figure}[h!]
    \centering
    \includegraphics[width=\linewidth]{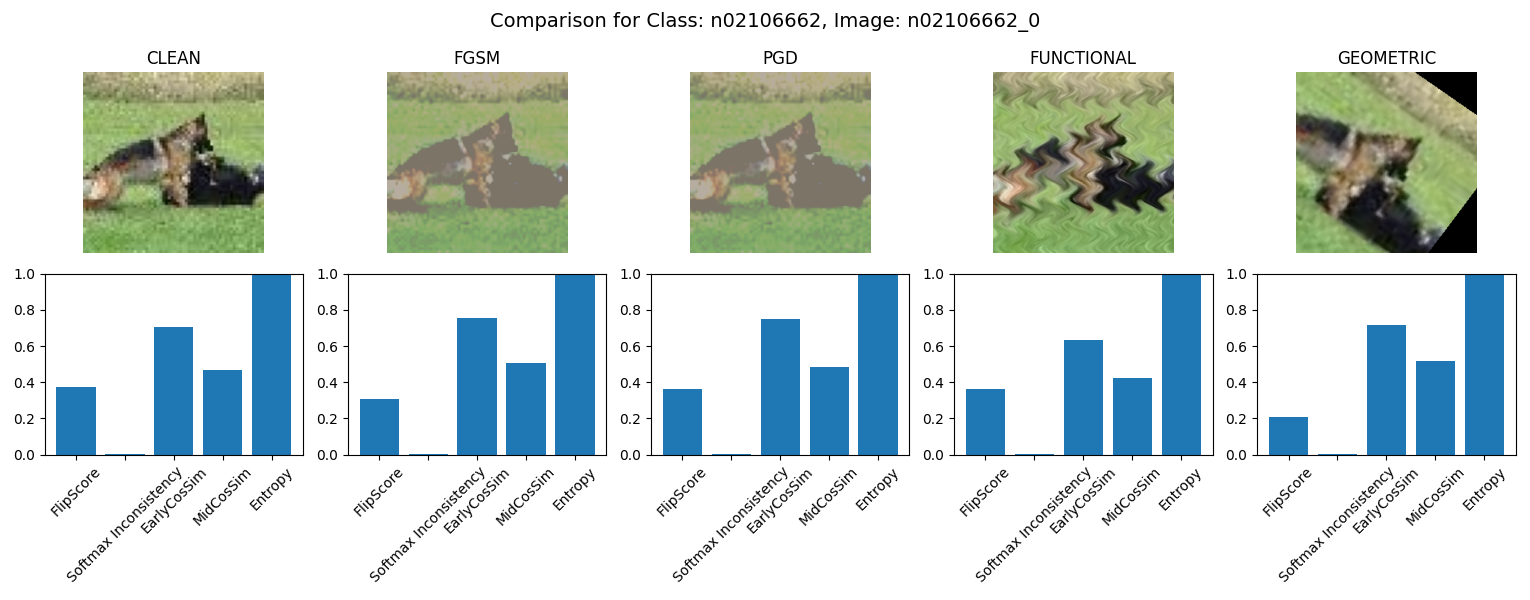}
    \caption{Qualitative comparison of instability metrics for clean and adversarial samples (example 2).}
    \label{fig:fig2_withoutpatch}
\end{figure}

\begin{figure}[h!]
    \centering
    \includegraphics[width=\linewidth]{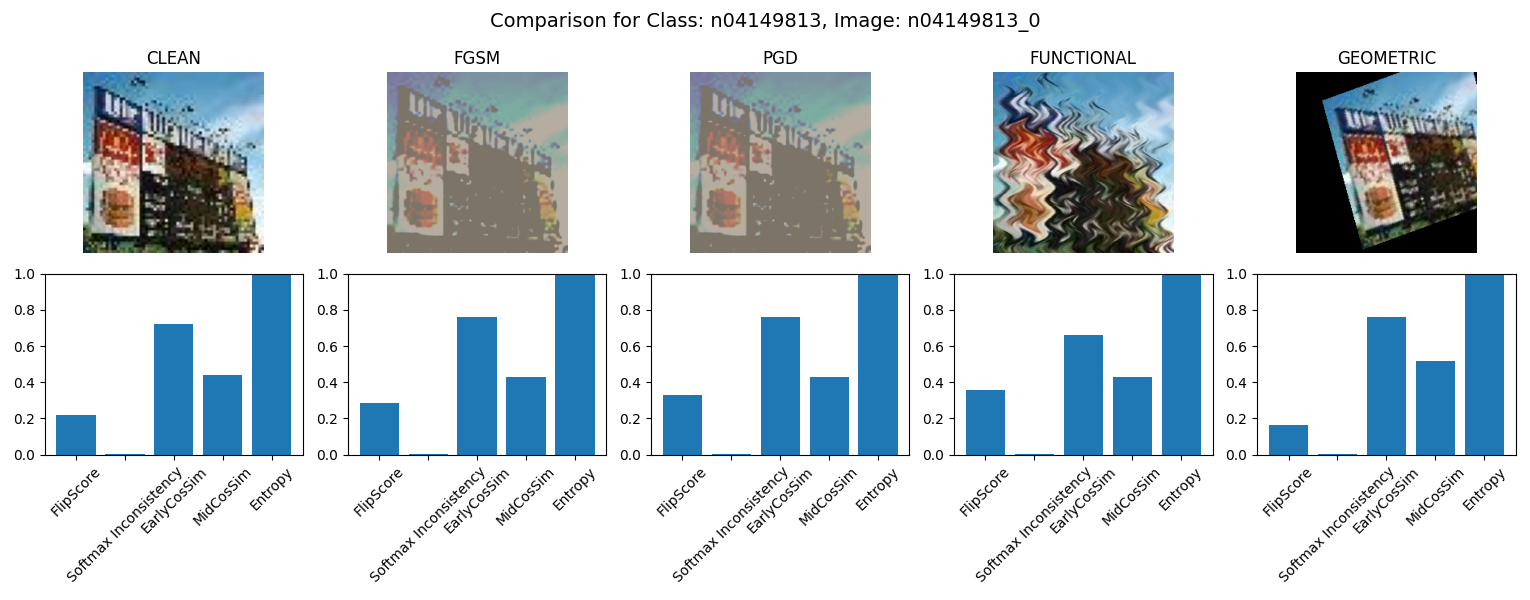}
    \caption{Qualitative comparison of instability metrics for clean and adversarial samples (example 3).}
    \label{fig:fig3_withoutpatch}
\end{figure}

To provide deeper insights into how different adversarial perturbations influence model interpretability, we present qualitative visualizations of instability metrics across representative examples from the Tiny ImageNet dataset (Figures~\ref{fig:fig1_withoutpatch}--\ref{fig:fig3_withoutpatch}).  For each input image, we show the clean version alongside its adversarial counterparts generated using FGSM, PGD, functional, and geometric attacks. Beneath each image, bar charts illustrate the five handcrafted metrics computed by LAFANet: \textit{FlipScore}, \textit{Inconsistency}, \textit{EarlyCosSim}, \textit{MidCosSim}, and \textit{Entropy}. These visualizations highlight how adversarial perturbations systematically increase prediction instability, reduce feature consistency, and amplify uncertainty compared to clean inputs.

\subsection{Interpretability and Real-World Scenario}

In AEGIS, interpretability is inherently provided through the five instability features computed by LAFANet, which are directly integrated into the decision-making pipeline. As illustrated in Figures~\ref{fig:fig1_withoutpatch}--\ref{fig:fig3_withoutpatch}, small benign transformations cause adversarial inputs to exhibit frequent label changes, feature misalignment, and increased entropy, while clean inputs remain stable under the same perturbations. These effects are consistently captured by FlipScore, Inconsistency, cosine similarity measures, and predictive entropy. For each sample, these indicators provide a concise and quantitative rationale, showing that adversarial examples systematically exhibit higher instability and uncertainty than clean samples. Because these same features are used both for classification and uncertainty-based rejection, the resulting explanations are inherently transparent and easily auditable.

A representative deployment scenario involves a low-speed autonomous shuttle operating in urban environments, where traffic signs may be manipulated using adversarial stickers or patch-based perturbations. In such settings, AEGIS first filters semantically inconsistent frames using SemantiGAN, then computes instability features and an evidential uncertainty score \(u\). If \(u > \tau_u\), the system rejects the input and activates a safe fallback strategy (e.g., slowing down or stopping) while logging diagnostic evidence such as increased inconsistency, reduced cosine similarity, and elevated entropy. Otherwise, the system proceeds with the predicted class. This design is well suited for real-time, resource-constrained edge deployment scenarios.

While adversarial examples may appear visually emphasized in the presented figures for interpretability purposes, in practice such perturbations are often imperceptible to human observers and can bypass common preprocessing techniques such as JPEG compression or smoothing. Therefore, dedicated detection pipelines such as AEGIS remain necessary to ensure robustness against modern adversarial threats.

\subsection{Security-Oriented Evaluations}

In safety-critical applications, the robustness of adversarial detection systems must be evaluated not only against known attack distributions but also against previously unseen and transferable adversarial threats. Accordingly, this subsection presents a comprehensive security assessment of AEGIS, including its generalization to zero-day adversarial examples, resistance to cross-model transfer attacks, and effectiveness in mitigating the threat scenarios defined in the proposed threat model.

\subsubsection{Zero-Day Generalization}

To evaluate the ability of AEGIS to detect previously unseen adversarial strategies, we simulate zero-day threats by generating hybrid perturbations that were not included during training. Specifically, we construct composite attacks combining PGD with geometric transformations such as random rotations and scaling. These hybrid adversarial samples are explicitly excluded from the training of both the SemantiGAN discriminator and the LAFANet module, ensuring a strict evaluation of out-of-distribution robustness. Detection is performed using the evidential classifier.

\begin{table*}[ht]
\centering
\caption{Detection performance on zero-day adversarial examples.}
\label{tab:zeroday}
\begin{tabular}{lccc}
\toprule
\textbf{Metric} & \textbf{Zero-Day AUROC (\%)} & \textbf{ECE (\%)} & \textbf{Entropy (↑)} \\
\midrule
Detection Performance & 86.5 & 4.9 & 1.48 \\
Confidence Drop (vs. Known Attacks) & -4.2 & +2.1 & +0.91 \\
\bottomrule
\end{tabular}
\end{table*}

As shown in Table~\ref{tab:zeroday}, AEGIS maintains strong detection performance under zero-day conditions, achieving an AUROC of 86.5\% while exhibiting increased entropy for novel perturbations. This behavior confirms that both the semantic discriminator and evidential classifier retain meaningful generalization capability even when attack distributions deviate from training conditions.

Previous evaluations on standard unseen attacks (e.g., AutoAttack, CW-L2, and DeepFool) reported AUROC values around 84--86\%, further supporting the robustness of the proposed pipeline under distribution shifts.

\subsubsection{Adversarial Transferability Resistance}

We further evaluate AEGIS under black-box transfer attacks, where adversarial examples generated on surrogate models are evaluated against the proposed pipeline. In this setting, adversarial samples are crafted using FGSM and PGD on external architectures (ResNet-50 and DenseNet-121), while AEGIS is trained on ResNet-18. This setup simulates realistic scenarios where an attacker does not have access to the target model.

\begin{table*}[ht]
\centering
\caption{Cross-model adversarial transfer detection performance.}
\label{tab:transferability}
\begin{tabular}{lcccc}
\toprule
\textbf{Source Model} & \textbf{Attack Type} & \textbf{Transfer Accuracy (\%)} & \textbf{AUROC (\%)} & \textbf{Entropy (↑)} \\
\midrule
ResNet-50    & FGSM & 89.4 & 91.2 & 1.25 \\
ResNet-50    & PGD  & 85.7 & 89.1 & 1.38 \\
DenseNet-121 & FGSM & 88.6 & 90.3 & 1.19 \\
DenseNet-121 & PGD  & 84.3 & 88.2 & 1.43 \\
\bottomrule
\end{tabular}
\end{table*}

The results in Table~\ref{tab:transferability} demonstrate that AEGIS exhibits strong resilience to transferable adversarial perturbations, with AUROC consistently above 88\% across all settings. In addition, entropy values remain elevated under transferred attacks, indicating that the evidential classifier effectively captures increased epistemic uncertainty in black-box scenarios. These findings confirm that the combination of semantic filtering and instability-driven evidential reasoning significantly reduces the effectiveness of transfer-based adversarial attacks.

\subsubsection{Threat Model Resolution Mapping}
\label{subsec:threat_resolution}

To evaluate the security coverage of AEGIS, we map each threat category to its corresponding empirical performance.
\begin{itemize}
    \item 
\textbf{$\ell_p$-bounded gradient-based attacks (FGSM, PGD):}  
These attacks are effectively mitigated through augmentation-driven instability profiling and evidential uncertainty estimation. High AUROC values for FGSM (96.1\%) and PGD (94.2\%) reported in Table~\ref{tab:quantitative_metrics} confirm strong separability between clean and adversarial samples. Furthermore, Table~\ref{tab:uncertainty_metrics} shows increased entropy and higher calibration error under attack, supporting reliable uncertainty-aware rejection.
\item 
\textbf{Localized spatial attacks (adversarial patches):}  
SemantiGAN provides effective semantic-level filtering, achieving 93.4\% precision and 89.6\% recall for adversarial detection (Table~\ref{tab:semantic_filtering}). This corresponds to a measurable improvement in downstream classification accuracy (+4.5\%), indicating strong mitigation of patch-based perturbations that disrupt localized regions while preserving global semantics.
\item 
\textbf{Semantic and functional attacks:}  
Functional perturbations are addressed through the combined effect of semantic discrimination and instability-based detection, achieving 89.4\% AUROC (Table~\ref{tab:quantitative_metrics}). Elevated entropy values in Table~\ref{tab:uncertainty_metrics} further confirm increased epistemic uncertainty under these transformations.
\item 
\textbf{Geometric transformations:}  
Geometric attacks represent the most challenging category, with a lower but still strong AUROC of 88.3\%. This indicates partial robustness and highlights the difficulty of handling spatially invariant transformations that preserve semantic content while altering geometry.
\item 
\textbf{Cross-model transfer attacks:}  
As shown in Table~\ref{tab:transferability}, AEGIS maintains transfer detection accuracy above 86.5\% and AUROC above 88.2\%, demonstrating robustness in black-box settings where adversarial examples are generated on surrogate architectures.
\item 
\textbf{Zero-day adversarial attacks:}  
Evaluation on hybrid PGD+geometric attacks (Table \ref{tab:zeroday}) yields an AUROC of 86.5\%, confirming generalization to unseen perturbation compositions. The observed confidence drop (- 4.2\%) relative to known attacks reflects increased difficulty under distribution shift.
\item 
\textbf{Evidential calibration behavior:}  
EDL-based uncertainty estimation consistently shows increased entropy and higher ECE under adversarial conditions (Table~\ref{tab:uncertainty_metrics}), enabling reliable detection of overconfident misclassifications and supporting safe rejection decisions.

\end{itemize}

\subsection{Comparison with State-of-the-Art}

To position AEGIS with respect to existing adversarial detection approaches, we perform a qualitative comparison with representative state-of-the-art methods spanning feature-space analysis, uncertainty estimation, semantic reasoning, and generative modeling. The selected baselines include classical detectors such as ODIN, LID, Mahalanobis Distance, MC Dropout, Defense-GAN, and DUQ, as well as recent semantic-aware and uncertainty-aware approaches including Semantic-Aware Adversarial Examples (SAE) \cite{zhang2024constructing}, SceneTAP \cite{cao2025scenetap}, and Conflict-Aware Evidential Deep Learning (C-EDL) \cite{barker2026robust}.

Table~\ref{tab:sota_comparison} presents a qualitative comparison of these methods according to several key characteristics, including semantic reasoning, feature-space modeling, instability analysis, uncertainty estimation, multi-class adversarial categorization, and real-time applicability. Existing approaches generally focus on a limited subset of these capabilities. Feature-space detectors such as LID and Mahalanobis Distance effectively identify distributional deviations but do not explicitly model semantic attack characteristics. Uncertainty-aware methods such as MC Dropout, DUQ, and C-EDL provide confidence estimates but lack dedicated mechanisms for semantic reasoning and transformation-driven robustness analysis. Similarly, semantic-aware approaches such as SAE and SceneTAP focus on contextual consistency and semantic manipulation detection but do not exploit instability signatures or calibrated uncertainty estimation.

In contrast, AEGIS integrates complementary detection principles within a unified architecture. SemantiGAN provides semantic adversarial categorization, LAFANet captures instability patterns under stochastic transformations, and Evidential Deep Learning delivers calibrated uncertainty estimates. Furthermore, unlike most existing approaches that formulate adversarial detection as a binary problem, AEGIS performs multi-class adversarial categorization across multiple attack families. As shown in Table~\ref{tab:sota_comparison}, AEGIS is the only framework that simultaneously combines semantic reasoning, feature-space analysis, instability profiling, uncertainty-aware learning, multi-class attack discrimination, and real-time deployment capability. This combination enables a more comprehensive and reliable adversarial detection framework capable of addressing both conventional perturbation-based attacks and emerging semantic adversarial manipulations.

\begin{table*}[ht]
\centering
\caption{Comparison with state-of-the-art}
\label{tab:sota_comparison}
\renewcommand{\arraystretch}{1.15}
\resizebox{14.5cm}{!}{\begin{tabular}{lcccccc}
\bottomrule
\textbf{Method} &
\textbf{Semantic} &
\textbf{Feature-Space} &
\textbf{Instability Analysis} &
\textbf{Uncertainty} &
\textbf{Multi-Class Detection} \\
\bottomrule

ODIN \cite{liang2018enhancing} & \ding{55} & \ding{55} &  \ding{55} &  \ding{55} &  \ding{55}\\
LID \cite{ma2018characterizing} &  \ding{55} &  \ding{51} &  \ding{55} &  \ding{55} &  \ding{55} \\
Mahalanobis \cite{lee2018simple} &  \ding{55} &  \ding{51} &  \ding{55} &  \ding{55} &  \ding{55}  \\
MC Dropout \cite{gal2016dropout} &  \ding{55} &  \ding{55} &  \ding{55} &  \ding{51} &  \ding{55}  \\
Defense-GAN \cite{samangouei2018defensegan} &  \ding{55} &  \ding{55} &  \ding{55} &  \ding{55} &  \ding{55}  \\
DUQ \cite{van2020uncertainty} &  \ding{55} &  \ding{51} &  \ding{55} &  \ding{51} &  \ding{55}  \\
SAE \cite{zhang2024constructing} &  \ding{51} &  \ding{55} &  \ding{55} &  \ding{55} &  \ding{55}  \\
SceneTAP \cite{cao2025scenetap} &  \ding{51} &  \ding{55} &  \ding{55} &  \ding{55} &  \ding{55}  \\
C-EDL \cite{barker2026robust} &  \ding{55} &  \ding{55} &  \ding{55} &  \ding{51} &  \ding{55}  \\
\bottomrule
\textbf{AEGIS} &  \ding{51} & \ding{51}  & \ding{51} & \ding{51} & \ding{51}   \\
\bottomrule
\end{tabular}
}
\end{table*}

\subsection{Ablation Study}

To quantify the contribution of each component in AEGIS, we perform an ablation study by selectively removing key modules, including the SemantiGAN discriminator, stochastic augmentation pipeline, handcrafted instability features, and the Evidential Deep Learning (EDL) classifier. The results are reported in Table~\ref{tab:ablation_study}.

The full AEGIS configuration achieves the best performance across all evaluation metrics, confirming the complementary role of its components. In particular, removing any individual module leads to a consistent degradation in both detection performance and calibration quality, demonstrating that each stage contributes meaningfully to the overall robustness of the system.

\begin{table*}[ht]
\centering
\caption{Ablation study of AEGIS components on detection and calibration performance.}
\label{tab:ablation_study}
  \resizebox{\textwidth}{!}{  

\begin{tabular}{lccccc}
\toprule
\textbf{Configuration} & \textbf{AUROC (\%)} & \textbf{AUPRC (\%)} & \textbf{Accuracy (\%)} & \textbf{ECE (\%)} & \textbf{Brier Score} \\
\midrule
\textbf{Full AEGIS Pipeline}     & \textbf{92.1} & \textbf{90.2} & \textbf{90.7} & \textbf{1.6} & \textbf{0.089} \\
w/o SemantiGAN                        & 87.5 & 84.6 & 86.2 & 3.9 & 0.104 \\
w/o Stochastic Augmentation           & 88.1 & 85.3 & 87.1 & 3.6 & 0.101 \\
w/o Instability Features              & 88.7 & 86.1 & 87.4 & 3.3 & 0.096 \\
Softmax instead of EDL                & 85.8 & 83.7 & 85.2 & 4.4 & 0.108 \\
No Defense (Baseline)                 & 80.2 & 78.5 & 82.3 & 5.7 & 0.117 \\
\bottomrule
\end{tabular}}
\end{table*}
\newpage
\subsection{Work Limitations}
\label{sec:limitations}

While AEGIS demonstrates strong robustness against a wide range of adversarial attacks, several limitations remain that highlight opportunities for future improvement.

\begin{itemize}
    \item \textbf{Geometric robustness:} Although performance on geometric attacks remains strong, it is comparatively lower than other attack categories. This suggests that explicitly geometry-invariant representations or equivariant feature learning could further improve robustness.

    \item \textbf{Zero-day generalization:} AEGIS maintains an AUROC above 86\% on zero-day adversarial examples; however, the observed reduction in confidence relative to known attacks (approximately 4\%) indicates reduced stability under highly novel perturbation distributions.

    \item \textbf{Real-time deployment:} All experiments were conducted in an offline setting. As such, latency, throughput, and computational efficiency under real-time constraints have not yet been evaluated, particularly on edge or embedded hardware.

    \item \textbf{Physical-world attacks:} The current evaluation does not include fully physical adversarial scenarios, such as printed patches, environmental perturbations, or real-world viewpoint distortions. Assessing robustness under such conditions remains an open challenge.

    \item \textbf{Scalability to higher resolutions:} While most experiments were conducted on Tiny ImageNet ($64 \times 64$), preliminary results on ImageNet-128 indicate that AEGIS generalizes well to higher resolutions, with only a minor degradation in performance (approximately 2\% AUROC drop). This suggests good scalability, although further evaluation on full-resolution datasets is required.
\end{itemize} 



The ablation analysis (Table~\ref{tab:ablation_study}) demonstrates that removing any individual component, SemantiGAN, LAFANet, instability feature construction, or EDL, results in a measurable degradation in detection accuracy, AUROC, and calibration quality. The full AEGIS pipeline achieves the best overall performance, confirming the complementary nature of its components.
Specifically:
\begin{itemize}
    \item \textbf{SemantiGAN} provides early-stage semantic filtering, reducing the propagation of high-risk inputs to downstream modules.
    \item \textbf{LAFANet} captures instability under stochastic transformations, revealing sensitivity patterns not visible in standard confidence scores.
    \item \textbf{Evidential Deep Learning (EDL)} enables calibrated prediction and explicit uncertainty quantification, supporting safe rejection under ambiguity.
\end{itemize}

Together, these components form a hierarchical defense strategy that integrates semantic reasoning, stability analysis, and uncertainty modeling into a unified adversarial detection framework.

\section{Conclusion }
\label{sec:conclusion}

This work presented AEGIS, a modular adversarial detection framework that integrates semantic filtering via a multi-class GAN discriminator, augmentation-driven instability profiling, and evidential deep learning for calibrated and uncertainty-aware classification. Extensive evaluations on Tiny ImageNet demonstrate that AEGIS consistently achieves high detection performance, strong calibration, and robust resilience against diverse adversarial threat models, including gradient-based, spatial, semantic, and transferable attacks.

The semantic discriminator enables early rejection of semantically inconsistent inputs, reducing the propagation of adversarial signals to downstream components. In parallel, LAFANet captures instability under stochastic transformations, providing discriminative features that reveal adversarial sensitivity beyond raw confidence scores. Finally, the evidential learning module enables principled uncertainty estimation, improving both robustness and interpretability of the final decision process.


While the current evaluation is conducted on controlled benchmarks, including Tiny ImageNet and ImageNet-128, the framework is designed to generalize to more complex real-world scenarios characterized by real time requirements, higher resolution, domain shift, and environment-induced variability. Future work will focus on optimizing low-latency inference for edge devices, extending the framework to privacy-preserving collaborative learning paradigms such as federated learning, and evaluating robustness under fully physical adversarial conditions, including printed and dynamic perturbations. Additional research directions include extending AEGIS to emerging adversarial attack categories and adapting the framework to real-time video analytics and multi-sensor vision systems.

\bibliographystyle{unsrt}  
\bibliography{ref}


\end{document}